\documentclass[runningheads]{llncs}
\usepackage[T1]{fontenc}
\usepackage{graphicx}
\usepackage{booktabs}
\usepackage{hyperref}
\usepackage{amsmath}
\usepackage{float} 

\usepackage[misc]{ifsym}

\usepackage{mwe}
\begin{document}

%
%

\title{Harnessing Deep LLM Participation for Robust Entity Linking}

\author{Jiajun Hou$^{\star }$ \and 
        Chenyu Zhang$^{\star}$ \and 
        Rui Meng\textsuperscript{(\Letter)}}
\authorrunning{J. Hou, C. Zhang, R. Meng}
\institute{Guangdong Provincial/Zhuhai Key Laboratory of IRADS, Beijing Normal-Hong Kong Baptist University, Zhuhai, China \\
\email{\{singularity738,zcydieyuki123\}@gmail.com, ruimeng@uic.edu.cn}\\
\renewcommand{\thefootnote}{\fnsymbol{footnote}}
\footnotetext[1]{Equal Contribution.}
}

\maketitle

\begin{abstract}

Entity Linking (EL), the task of mapping textual entity mentions to their corresponding entries in knowledge bases, constitutes a fundamental component of natural language understanding. Recent advancements in Large Language Models (LLMs) have demonstrated remarkable potential for enhancing EL performance. Prior research has leveraged LLMs to improve entity disambiguation and input representation, yielding significant gains in accuracy and robustness. However, these approaches typically apply LLMs to isolated stages of the EL task, failing to fully integrate their capabilities throughout the entire process.

In this work, we introduce DeepEL, a comprehensive framework that incorporates LLMs into every stage of the entity linking task. Furthermore, we identify that disambiguating entities in isolation is insufficient for optimal performance. To address this limitation, we propose a novel self-validation mechanism that utilizes global contextual information, enabling LLMs to rectify their own predictions and better recognize cohesive relationships among entities within the same sentence.

Extensive empirical evaluation across ten benchmark datasets demonstrates that DeepEL substantially outperforms existing state-of-the-art methods, achieving an average improvement of 2.6\% in overall F1 score and a remarkable 4\% gain on out-of-domain datasets. These results underscore the efficacy of deep LLM integration in advancing the state-of-the-art in entity linking.
\end{abstract}

\section{Introduction}

With the proliferation of Internet technology, an unprecedented volume of natural language data is being generated across social media platforms and communication channels. This exponential growth has significantly accelerated research in natural language processing (NLP)~\cite{guellil2024entity}. Entity Linking (EL), the process of accurately mapping entity mentions in text to corresponding entries in knowledge bases, has become increasingly crucial across various applications, including Question Answering~\cite{pan2019improving,baek2023knowledge,yao2023korc}, Information Retrieval~\cite{hambarde2023information}, and Machine Translation~\cite{babych2003improving}.

The inherent complexity of EL stems from several linguistic phenomena present in natural languages. First, synonymy presents the challenge of multiple surface forms referring to the same entity (e.g., ``IBM'' and ``International Business Machines'' denoting the same organization). Second, homonymy occurs when identical terms represent different entities (e.g., ``Apple'' referring to either a fruit or a technology company). These linguistic challenges necessitate entity linking models with robust contextual understanding and strong generalization capabilities. Consequently, research in entity linking has evolved from early dictionary-based approaches to current methods leveraging deep learning and pre-trained models~\cite{van2020rel,wu2019scalable,ayoola2022refined}. However, although deep learning models generally outperform dictionary-based approaches in generalization, their performance on both seen and unseen datasets remains inconsistent due to limitations in training data volume and parameter capacity. When encountering entities absent from the training data, their performance deteriorates significantly.

The emergence of Large Language Models (LLMs), built upon multi-head attention mechanisms~\cite{ashish2017attention}, offers a promising solution to these challenges. LLMs have demonstrated exceptional performance across various natural language processing tasks, prompting recent investigations into their application for entity linking. Notably, ChatEL~\cite{ding2024ChatEL} and LLMAEL~\cite{xin2024llmael} represent significant advancements in this domain. ChatEL employs LLMs to perform entity disambiguation as a multiple-choice task, while LLMAEL focuses on input augmentation by generating entity descriptions with LLMs. Although these approaches successfully incorporate LLMs into the entity linking task and achieve competitive results on standard benchmarks, their integration of LLMs remains limited to specific components rather than encompassing the entire entity linking process.

Given the remarkable generalization capabilities of LLMs for entity linking, we advocate for their comprehensive integration throughout the EL task. To this end, we propose the \textbf{DeepEL} framework. First, in the candidate entity generation phase, traditional models are often constrained by insufficient contextual information to produce high-quality results. DeepEL leverages LLMs to initially describe the entities requiring linkage, providing richer contextual information that enables pre-trained models to generate more accurate candidate lists. Subsequently, DeepEL formulates entity disambiguation as a multi-selection task for LLMs using the generated candidate list and the descriptions of candidate entities from the knowledge base. The aforementioned two steps disambiguate each entity mention in isolation, mapping them to knowledge base candidates while disregarding the cohesive relationships between entities within the same context. This oversight can lead to suboptimal linking outcomes. DeepEL innovatively addresses this limitation by incorporating a self-validation mechanism. Following initial entity disambiguation, it cross-validates each entity against the linking results of all other entities in the sentence. Through this global consistency check utilizing inter-entity context, DeepEL significantly enhances the overall accuracy of entity linking. By applying LLMs to every stage of the entity linking process and employing them for self-validation, DeepEL achieves truly deep LLM integration.

Our comprehensive experimental evaluation provides extensive comparisons with state-of-the-art entity linking models. The results demonstrate that DeepEL achieves substantial improvements, with an average 2.6\% increase in F1 score across all datasets and a notable average 4\% improvement on out-of-domain datasets, validating the effective utilization of LLMs' generalization capabilities.

The main contributions of this work are summarized as follows:

\begin{itemize}
    \item We introduce DeepEL, an entity linking framework powered by Large Language Models (LLMs) that leverages LLMs at every stage—candidate generation, entity disambiguation, and self-validation—to thoroughly harness their strengths and enhance entity linking effectiveness.
    \item We propose an approach to assist pre-trained models in generating higher-quality entity candidate lists by employing LLMs to provide descriptive context for entities requiring linkage.
    \item We introduce a novel self-validation mechanism based on global contextual information, which improves the accuracy of entity disambiguation by providing LLMs with information about other entities within the same context.
    \item We conduct comprehensive comparisons with five state-of-the-art models across ten benchmark datasets, demonstrating DeepEL's superior performance. 
    We have made our code publicly available~\footnote{https://github.com/SStan1/DeepEL} to guarantee reproducibility.
\end{itemize}

\section{Related Work}
Entity Linking (EL) aims to map entity mentions in text to corresponding entries in a knowledge base. Early EL systems relied heavily on surface form dictionaries, string similarity, and heuristic rules constructed from large-scale resources such as Wikipedia~\cite{bunescu-pasca-2006-using,cucerzan2007large}. These methods typically matched mentions to candidate entities based on surface form overlap and simple context heuristics, which limited their ability to handle ambiguity and context-dependent cases.

To address these challenges, Milne and Witten~\cite{milne2008learning} proposed a method that leverages Wikipedia's anchor links to compute statistical relatedness between entities, greatly improving disambiguation accuracy and establishing a widely used baseline. Ratinov et al.~\cite{ratinov2011local} systematically explored both local and global disambiguation algorithms, demonstrating that combining mention-level features with document-level coherence can further boost performance. Hoffart et al.~\cite{hoffart2011robust} introduced the AIDA system, which formulates EL as a global optimization problem by maximizing the overall coherence of all entity assignments within a document. This approach set a classic benchmark and inspired many subsequent works.

With the rapid development of deep learning, neural models have become the mainstream approach for EL. BLINK~\cite{wu2019zero} employs a bi-encoder architecture to efficiently retrieve and rank candidate entities, encoding mention contexts and candidate entities separately and using vector similarity for fast retrieval. REL~\cite{van_Hulst_2020} advances this by jointly modeling mention context and entity relationships, combining contextual and relational features to achieve state-of-the-art results on several public benchmarks. For zero-shot scenarios, Refined~\cite{ayoola2022refined} proposes an efficient end-to-end approach that integrates contextualized representations and efficient inference, achieving strong generalization to unseen entities.

In addition to traditional classification and ranking frameworks, recent studies have explored alternative task formulations. EntityQA~\cite{zhang2021entqa} reframes EL as a question answering problem, where the system generates answers to implicit questions about entity mentions, which is particularly effective for ambiguous cases. GENRE~\cite{de2020autoregressive} treats EL as a sequence generation task, leveraging encoder-decoder architectures to directly generate entity names token by token, showing advantages in multilingual and open-domain settings.

With the advent of large language models, a new research direction has emerged: applying large language models to entity linking tasks. In this context, two recent works are particularly noteworthy: ChatEL~\cite{ding2024ChatEL} and LLMAEL~\cite{xin2024llmael}. ChatEL proposes that a pre-trained entity linking model can be used to generate a list of candidate entities, which are then presented as a multiple-choice question for LLMs to perform entity disambiguation. In contrast, LLMAEL utilizes LLM-generated descriptions of entities as enhanced information to assist the entity linking model. Both works have achieved excellent results on public datasets, demonstrating the extraordinary potential of LLMs in entity linking.

\section{Problem Definition}
Entity Linking (EL) is the task of mapping mentions in text to their corresponding entities in a knowledge base. We formally define the problem as follows: Let \( KB \) denote a knowledge base containing a set of entities \( \{e_1, e_2, \dots, e_n\} \), where each \( e_i \) represents a unique real-world object. Given an input text \( T \) containing multiple entity mentions \( M = \{m_1, m_2, \dots, m_k\} \), the text can be represented as \( c = \dots \, || \, t_1 \, || \, m_1 \, || \, t_2 \, || \, m_2 \, || \, \dots \), where \( t_i \) denotes text spans and \( m_i \) denotes entity mentions. The objective of entity linking is to determine the correct mapping from each mention \( m_i \) to its corresponding entity \( e_i \in KB \), resulting in a set of mention-entity pairs \( \{(m_i, e_i)\}_{i=1}^k \).

\section{Methodology}
DeepEL, our proposed approach for integrating LLM into the entity linking task, comprises three primary modules:

\begin{enumerate}
  \item \textbf{Entity Candidate Generation:}  Since large language models cannot directly connect to databases, we will first let the LLM describe the entities, and then let the pre-trained entity linking model generate a high-quality list of sentence entities based on the LLM's descriptions and the original sentences.
  \item \textbf{Entity Disambiguation:} Based on the entity list that has been generated, we organize the entity disambiguation task as a multiple choice question for LLMs to perform entity disambiguation.
  \item \textbf{Self-Validation:}  Traditional entity disambiguation methods tend to disambiguate entities in isolation. To address this limitation and better consider contextual information, we input information about other entities in the same sentence into the LLM as global information and ask it to self-validate its previous judgments based on this information.The LLMs will re-select the negatively judged entities.
\end{enumerate}

\begin{figure}[H]
    \centering
    \vspace{-10pt}  
    \includegraphics[height=\textwidth]{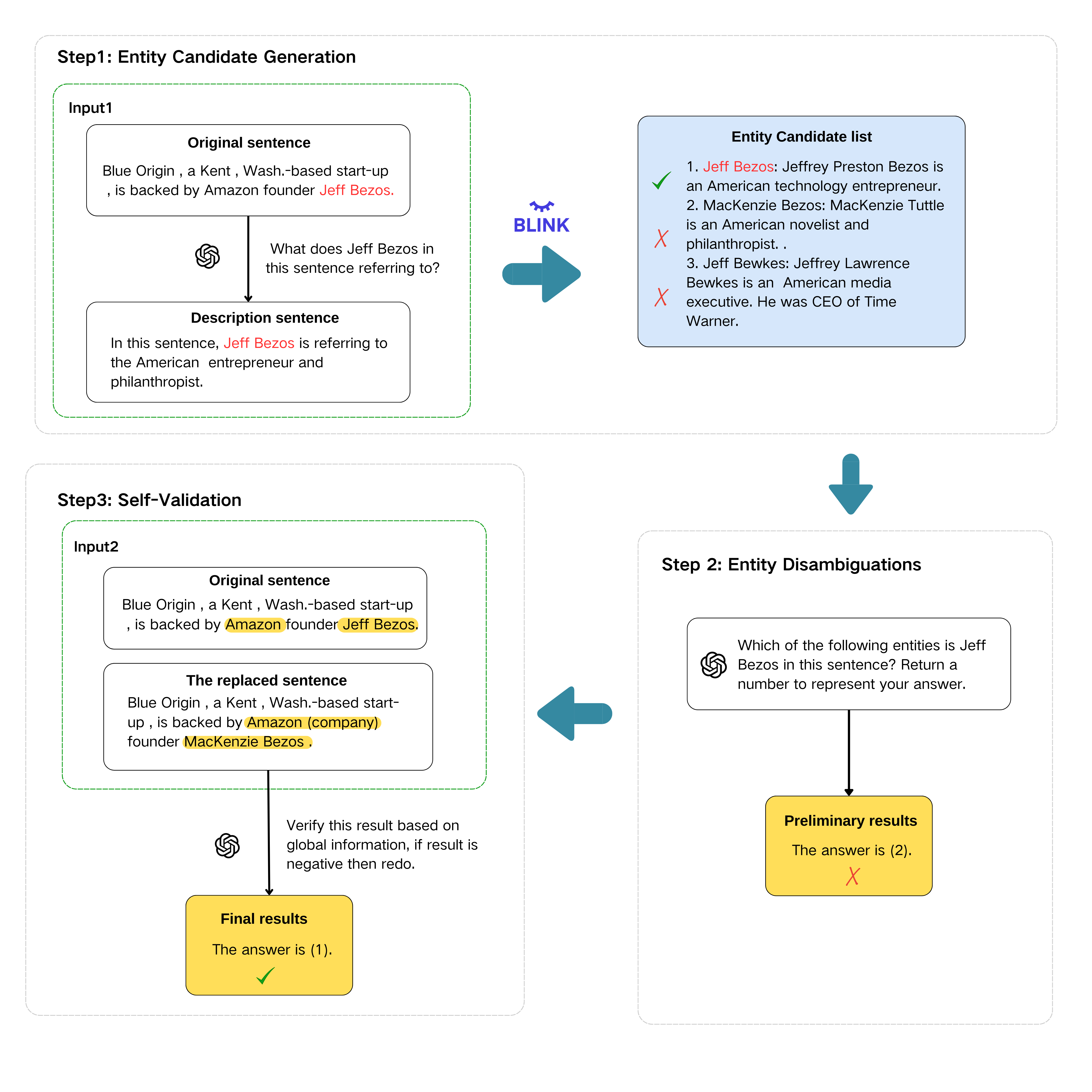}
    \vspace{-10pt}  
    \caption{DeepEL's workflow. Text marked in red indicates the target entity mention currently being linked. Highlighted text represents already-linked entity mentions within the same sentence, providing contextual linking information for the target entity.}
    \label{fig:example}
    \vspace{-5pt}  
\end{figure}

\subsection{Entity Candidate Generation}

Large language models (LLMs) lack direct access to structured knowledge bases, creating a fundamental challenge for entity linking tasks. To address this limitation, we develop a novel hybrid approach that combines the interpretative capabilities of LLMs with the retrieval efficiency of traditional entity linking systems.

Given an input text \(T\) containing entity mentions \(M = \{m_1, m_2, \dots, m_k\}\), we implement a two-stage candidate generation process that leverages both LLM-based interpretation and traditional retrieval methods:

\begin{enumerate}
    \item \textbf{LLM Interpretation}: We prompt the LLM to analyze the input text and generate a contextual interpretation \(I_{m_i}\) for each mention \(m_i \in M\). These interpretations capture the LLM's understanding of the entities based on contextual cues and its parametric knowledge.
    
    \item \textbf{Dual-Source Candidate Retrieval}: We use BLINK to generate two complementary candidate lists. The dense retrieval used by BLINK can generate more options to guarantee the diversity of results.
    \begin{itemize}
        \item \(C_{original}\): candidates retrieved using the original text \(T\) as context
        \item \(C_{LLM}\): candidates retrieved using the LLM's interpretations \(\{I_{m_1}, I_{m_2}, \dots, I_{m_k}\}\) as context
    \end{itemize}
\end{enumerate}

Each candidate list is constrained to the top ten entities to maintain computational efficiency. For each candidate entity \(e\), we retrieve its canonical description from the knowledge base \(KB\), forming entity-description pairs that provide essential information for disambiguation:

\begin{equation}
C_{original} = \{(e_1, d_1), \dots, (e_n, d_n)\} \text{ where } d_i = KB.description(e_i)
\end{equation}
\begin{equation}
C_{LLM} = \{(e'_1, d'_1), \dots, (e'_m, d'_m)\} \text{ where } d'_i = KB.description(e'_i)
\end{equation}

To optimize both coverage and efficiency, we merge these candidate lists using a priority-based algorithm that ensures the final list remains manageable while incorporating the strengths of both retrieval methods:

\begin{equation}
C_{final} = f_{merge}(C_{original}, C_{LLM}) \text{ such that } |C_{final}| \leq 10
\end{equation}

This dual-source candidate generation approach significantly enhances the quality and diversity of candidate entities. By combining traditional context-based retrieval with LLM interpretations, our method captures both explicit textual cues and implicit semantic connections that might be missed by either approach alone. Each element in \(C_{final}\) preserves the entity-description pair structure, providing rich knowledge base information that is crucial for the subsequent disambiguation and verification steps.

\subsection{Entity Disambiguation}

This step utilizes the previously defined elements: the original mention \(m_i\), the LLM's interpretation \(I_{m_i}\), and the merged candidate list \(C_{final}\). Following Ding et al. (2024), for each mention \(m_i\), we formulate the entity disambiguation task as a multiple-choice problem:

\begin{equation}
f_{select}(m_i, I_{m_i}, C_{final}) \rightarrow (e_j, d_j), \text{ where } (e_j, d_j) \in C_{final} \cup \{\emptyset\}
\end{equation}
,where \(\emptyset\) indicates that none of the candidates in \(C_{final}\) is correct. To ensure unambiguous responses, we require the LLM to output the index \(j\) of its chosen entity from \(C_{final}\), with 0 representing \(\emptyset\). This transforms the entity selection into a well-defined choice based on both the mention, its interpretation, and the available candidate entities from the knowledge base.

\subsection{Self-Validation}

To enhance entity linking accuracy through contextual coherence, we implement a self-validation mechanism that leverages global information from the entire sentence context. This approach addresses a critical limitation in traditional entity linking methods, which often disambiguate each entity mention in isolation without considering the cohesive relationships between entities appearing in the same context.

Given our original text \(T\) containing \(k\) entity mentions \(M = \{m_1, ..., m_k\}\) and their initially predicted entities \(E = \{e_1, ..., e_k\}\) (selected through the entity disambiguation function \(f_{select}\)), we construct comprehensive global information for validation as follows:

First, we generate an entity-replaced text by substituting each mention with its predicted entity:
\begin{equation}
T_g = T[m_1 \rightarrow e_1, m_2 \rightarrow e_2, ..., m_k \rightarrow e_k]
\end{equation}

The descriptions \(D = \{d_1, ..., d_k\}\) for all predicted entities, which were previously retrieved from the knowledge base \(KB\) during the candidate generation phase, provide essential semantic information about each entity.

The global context \(G\) used for validation integrates three key components:
\begin{equation}
G = (T, T_g, D)
\end{equation}

This rich contextual representation enables the LLM to cross-validate each entity against the linking results of all other entities in the sentence. We prompt the LLM to perform self-validation based on this global information, requesting concise explanations alongside its validation decisions. For entities that fail validation (i.e., where the LLM determines the initial linking is inconsistent with the global context), we re-prompt the model with the same global context to perform entity re-selection from the original candidate list.

This self-validation process significantly enhances linking accuracy by ensuring global consistency across all entity mentions within the same context. By examining the inter-entity relationships and contextual coherence, the model can correct initial disambiguation errors that might occur when entities are processed in isolation. For computational efficiency, this self-validation process is performed only once, and re-evaluated entity assignments are not subjected to further validation rounds.

\section{Experiment}
In this section, we comprehensively compare DeepEL with several state-of-the-art supervised models across ten benchmark datasets. We further analyze the quality of entity candidate lists generated in Step 1 and evaluate the effectiveness of the self-validation method in Step 3. Finally, we conduct case-specific error analyses to identify potential improvements to our framework.

\subsection{Experimental Setup}

\subsubsection{Datasets}
Following ChatEL~\cite{ding2024ChatEL}, our experiments utilize eight standard datasets from their work, along with two additional commonly used datasets. By convention, we categorize these datasets into in-domain and out-of-domain groups for baseline evaluation:

Out-of-Domain: OKE15~\cite{nuzzolese2015open}, OKE16~\cite{nuzzolese2015open}, DER~\cite{derczynski2015analysis}, KORE~\cite{hoffart2012kore}, REU, and RSS~\cite{roder2014n3}. These datasets assess a model's ability to generalize to new domains and adapt to unfamiliar contexts.
In-Domain: ACE04~\cite{ratinov2011local}, AIDA~\cite{hoffart2011robust}, MSN~\cite{cucerzan2007large}, and AQU~\cite{milne2008learning}. These datasets evaluate model performance within familiar contexts.

\subsubsection{Baselines}
We compare DeepEL against the following baselines:

\begin{itemize}
    \item \textbf{REL}~\cite{van2020rel}: An entity linking model that combines deep learning with traditional retrieval techniques. It leverages pre-trained language models and precomputed indices for efficient candidate generation, improving both precision and speed.
    \item \textbf{BLINK}~\cite{wu2019zero}: A two-stage entity linking framework utilizing pre-trained BERT models for accurate disambiguation. BLINK is a core component of DeepEL, making this comparison particularly important.
    \item \textbf{GENRE}~\cite{de2020autoregressive}: This approach reframes entity disambiguation as an entity name generation task, employing generative models (e.g., BART) to directly produce the target entity's name instead of relying on candidate generation and ranking.
    \item \textbf{ReFinED}~\cite{ayoola2022refined}: A recent system from Amazon that leverages Wikipedia entries as additional features to enhance entity disambiguation and linking, particularly improving performance on ambiguous and long-tail entities.
    \item \textbf{ChatEL}~\cite{ding2024ChatEL}: An LLM-based entity linking method that partially inspired our approach. Some evaluation results are taken directly from the ChatEL paper. Notably, to avoid the evaluation errors discussed previously, we modified ChatEL's prompt to output numerical choices, resulting in significantly different results compared to the original paper.
\end{itemize}

\subsubsection{Evaluation Metric}
To ensure consistency and comparability across both in-domain and out-of-domain datasets and all five models, we adopt the evaluation protocol from ChatEL~\cite{ding2024ChatEL}. Specifically, we report the in-KB micro-F1 score, which excludes entities not present in the knowledge base from evaluation.

\subsection{Main Result}
\begin{table*}[ht]
\centering
\renewcommand{\arraystretch}{1.5}
\caption{Shows the F1 score of DeepEL on ten standard datasets, where avg-OOD represents the average of the model's performance on the out-domain dataset. The results of the best performing model for each of these datasets are bolded and the second best are underlined.}
\label{tab:f1-score-deepel}
\resizebox{\textwidth}{!}{%
\begin{tabular}{|c|cccccc|cccc|cc|}
\hline
\textbf{} & \multicolumn{6}{c|}{\textbf{Out-of-domain}} & \multicolumn{4}{c|}{\textbf{In-domain}} & \textbf{} & \textbf{} \\
\hline
\textbf{} & \textbf{KORE} & \textbf{OKE15} & \textbf{REU} & \textbf{RSS} & \textbf{DER} & \textbf{OKE16} & \textbf{AQU} & \textbf{ACE04} & \textbf{MSN} & \textbf{AIDA} & \textbf{Average} & \textbf{avg-OOD} \\
\hline
\textbf{REL} & 0.618 & 0.705 & 0.662 & 0.680 & 0.411 & 0.749 & \textbf{0.881} & \textbf{0.897} & \textbf{0.930} & 0.805 & 0.7338 & 0.6375 \\
\textbf{GENRE} & 0.542 & 0.640 & 0.697 & 0.708 & 0.541 & 0.708 & 0.849 & 0.848 & 0.780 & \underline{0.837} & 0.7150 & 0.6393 \\
\textbf{ReFinED} & 0.567 & \underline{0.781} & 0.680 & 0.708 & 0.507 & \underline{0.794} & 0.861 & 0.864 & 0.891 & \textbf{0.840} & 0.7493 & 0.6728 \\
\textbf{BLINK} & 0.618 & 0.763 & 0.712 & 0.767 & 0.709 & \textbf{0.805} & \underline{0.863} & 0.793 & 0.865 & 0.807 & 0.7702 & 0.7290 \\
\textbf{ChatEL} & \underline{0.787} & 0.758 & \underline{0.789} & \underline{0.822} & \underline{0.717} & 0.752 & 0.767 & \underline{0.893} & 0.881 & 0.821 & \underline{0.7987} & \underline{0.7708} \\
\textbf{DeepEL} & \textbf{0.873} & \textbf{0.833} & \textbf{0.812} & \textbf{0.857} & \textbf{0.768} & 0.762 & 0.753 & 0.888 & \underline{0.905} & 0.799 & \textbf{0.8250} & \textbf{0.8175} \\
\hline
\end{tabular}%
}
\end{table*}

Our systematic experimental evaluation compared DeepEL with five state-of-the-art entity linking models across multiple datasets. As illustrated in Table 1, DeepEL demonstrates exceptional overall performance, surpassing ChatEL by 2.6\% in average F1 score, which strongly validates the effectiveness of our methodological framework. 

The most significant finding emerges in DeepEL's remarkable generalization capability. On out-of-domain datasets, DeepEL achieves a substantial 4\% improvement in average F1 score compared to ChatEL. This result provides compelling evidence that our approach successfully leverages the advanced inference and contextual understanding capabilities of large language models, enabling robust entity linking performance even when confronted with previously unseen data distributions.

Despite DeepEL's impressive generalization capabilities, we observe a more nuanced performance pattern on in-domain datasets. Across the four in-domain benchmarks, DeepEL does not consistently achieve optimal results compared to specialized models. We attribute this performance characteristic primarily to the inherent inference instability of large language models (LLMs). 

Although we implemented a self-validation mechanism specifically designed to enhance model stability, DeepEL still faces challenges in outperforming traditional deep learning methods that are meticulously optimized for particular domains. Interestingly, ChatEL—which also employs LLMs in its architecture—encounters similar challenges but achieves more consistent in-domain performance, likely due to its reduced dependency on LLM components. This comparative analysis further substantiates our hypothesis that the inherent variability in LLM inference processes can adversely affect the precision of entity linking tasks in well-established domains where specialized models have been extensively fine-tuned.

\subsubsection{Different LLM Models in DeepEL}

\begin{table*}[ht]
\centering
\renewcommand{\arraystretch}{1.5}
\caption{Performance of DeepEL using different LLM models}
\label{tab:Ablation Study}
\resizebox{\textwidth}{!}{%
\begin{tabular}{|c|cccccc|cccc|c|}
\hline
\textbf{} & \multicolumn{6}{c|}{\textbf{Out-of-domain}} & \multicolumn{4}{c|}{\textbf{In-domain}} & \textbf{}  \\
\hline
\textbf{} & \textbf{KORE} & \textbf{OKE15} & \textbf{REU} & \textbf{RSS} & \textbf{DER} & \textbf{OKE16} & \textbf{AQU} & \textbf{ACE04} & \textbf{MSN} & \textbf{AIDA} & \textbf{Average} \\
\hline
\textbf{llama-2-70b} & 0.814 & 0.708 & 0.650      & 0.789      & 0.678       & 0.601     & 0.563   & 0.753       & 0.788        & 0.678        & 0.7022           \\

\textbf{GPT 3.5} & 0.769 & 0.708 & 0.712       & 0.803       & 0.718        & 0.619        & 0.634        & 0.773        & 0.847        & 0.634        & 0.7302           \\

\textbf{GPT 4}         & \textbf{0.873}         & \textbf{0.833}         & \textbf{0.812} & \textbf{0.857} & \textbf{0.768}        & \textbf{0.762}        & \textbf{0.753} & \textbf{0.888} &  \textbf{0.905} & \textbf{0.799}     & \textbf{0.8250} \\
\hline
\end{tabular}%
}
\end{table*}

We evaluated DeepEL using various large language model (LLM) backbones, including GPT-3.5, GPT-4, and Llama-2 (with versions above 3). The results, presented in Table 2, yield several key observations. 

First, the scores achieved by the three models exhibit a clear gradient corresponding to their respective capabilities, with GPT-4 achieving the highest scores and Llama-2 the lowest. This pattern demonstrates that DeepEL functions as an adaptable framework, with performance scaling alongside the capabilities of the underlying LLMs—a property that endows DeepEL with significant potential for future applications. 

Second, the relatively close scores between GPT-3.5 and Llama-2, despite their architectural differences, highlight DeepEL's generalizability across diverse model families. This finding suggests that DeepEL can be effectively deployed with a wide range of LLMs while consistently enhancing their entity linking performance

\subsection{Ablation Study}
\begin{table*}[ht]
\centering
\renewcommand{\arraystretch}{1.5}
\caption{Ablation study of Entity descriptions and Global Self-validation}
\label{tab:Ablation Study}
\resizebox{\textwidth}{!}{%
\begin{tabular}{|c|cccccc|cccc|c|}
\hline
\textbf{} & \multicolumn{6}{c|}{\textbf{Out-of-domain}} & \multicolumn{4}{c|}{\textbf{In-domain}} & \textbf{}  \\
\hline
\textbf{} & \textbf{KORE} & \textbf{OKE15} & \textbf{REU} & \textbf{RSS} & \textbf{DER} & \textbf{OKE16} & \textbf{AQU} & \textbf{ACE04} & \textbf{MSN} & \textbf{AIDA} & \textbf{Average} \\
\hline
\textbf{DeepEL}         & \textbf{0.873}         & \textbf{0.833}         & \textbf{0.812} & \textbf{0.857} & \textbf{0.768}        & \textbf{0.762}        & \textbf{0.753} & \textbf{0.888} &  0.905 & 0.799      & \textbf{0.8250} \\

\textbf{DeepEL W/O De} & 0.806 & 0.790 & 0.796       & 0.841       & 0.751        & 0.734        & 0.681        & 0.887        & 0.898       & \textbf{0.806}        & 0.7990            \\

\textbf{DeepEL W/O Val} & 0.870 & 0.820 & 0.794       & 0.850       & 0.762        & 0.742        & 0.744        & 0.869        & \textbf{0.906}      & 0.790        & 0.8147            \\
\hline
\end{tabular}%
}
\end{table*}
We conduct ablation studies to further demonstrate the contribution of individual model components. Our main focus is to explore (1) whether adopting the Entity Description strategy improves performance; and (2) whether the Global Self-validation module helps with error detection. Therefore, we first test direct candidate set generation without allowing LLM rewrites, denoted as "W/O De". Moreover, we also test our model without using global self-validation, denoted as "W/O Val".

We performed ablation studies across ten benchmarks. For entity descriptions, we achieved the research goal by eliminating this step and directly letting BLINK generate a list of candidate entities for the original sentence. As shown in Table 3, removing the entity descriptions step impairs DeepEL's performance across all benchmark datasets. This demonstrates that having LLMs provide information about corresponding entities effectively enhances the connection between mentioned entities and target entities, thus improving the quality of the generated entity list.

As for Step 3, we similarly eliminated the self-validation step and directly used the initial selection as the final result. By comparing the data in the table, we conclude that the self-validation step consistently improves the performance of the DeepEL framework. This is because self-checking answer correctness through global information and making new choices accordingly effectively reduces misleading errors.

We also observed that on both AIDA and MSN datasets, the results obtained without Step 1 are actually higher. This occurs because AIDA and MSN, as in-domain datasets, are heavily trained, making the candidate entities generated by BLINK more accurate and stable in comparison, which enables LLMs to more easily approach the correct answers when making responses.

\section{Error Analysis And Case Study}

\subsection{Error Analysis}

\begin{table*}[ht]
\centering
\renewcommand{\arraystretch}{1.5}
\caption{Examples of DeepEL errors}
\label{tab:Error Analysis}
\resizebox{\textwidth}{!}{%
\begin{tabular}{|c|c|c|c|c|}
\hline

\textbf{Datasets} & \textbf{Entity Name} & \textbf{Ground Truth} & \textbf{Prediction} & \textbf{Error Types} \\
\hline
KORE & Stefani & Lady Gaga    & Gwen Stefani  & GT is incorrect  \\ 
KORE & Desire (Bob Dylan album) & Desire (Bob Dylan album) &   & Not give answer \\ 
KORE & Sony & Sony    &Sony Music  & Narrow  \\ 
ACE04 &  Moscow         &  Moscow&Russia&Expand\\
ACE04 &  White House  &  White House & White House Office&Narrow\\ 
ACE04 &  Volvo  &  Volvo & Volvo Cars & Narrow\\ 
RSS &  Apple Inc.  &  Steve Jobs & Apple Inc. & GT is incorrect\\
RSS &  NASA  &  Jet Propulsion Laboratory & NASA & GT is incorrect\\
RSS &  United States Navy  &  United States Navy & Navy & Expand\\
RSS &  Washington, D.C.  &  Washington, D.C. & & Not give answer\\
REU & Volkswagen Group  & Volkswagen Group & Volkswagen& Expand\\
AQU &  Civil war  &  Civil war & & Not give answer\\
AQU &   Pacific Northwest &   Pacific Northwest &  Northwestern United States& Narrow\\
AQU &  Wind  &  Wind &Wind gust & Narrow\\
MSN &  Pat Riley  &  Pat Riley & & Not give answer\\
MSN &  MVP  &  Most valuable player &  NBA Most Valuable Player Award& Narrow\\
OKE15 &  France  &  France &France during World War II & Narrow\\
OKE15 &  KwaZulu Provincial Division  &  KwaZulu-Natal Division & & Not give answer\\
OKE15 &  Salvatore Quasimodo  &  Salvatore Quasimodo &Quasimodo & Expand\\
OKE16 &  Country  &  Country &Italy & Narrow\\

\hline
\end{tabular}%
}
\end{table*}

We analyzed the records of error occurrences and identified four main categories: (1) incorrect ground truth; (2) expanding the range of answers; (3) narrowing the range of answers; and (4) not providing an answer.

The first type, denoted as \textit{GT is incorrect}, occurs when DeepEL selects the correct answer but is judged as incorrect due to flawed ground truth. For example, in the KORE dataset in Table 4, for the entity "Stefani", "Gwen Stefani" is clearly more accurate than "Lady Gaga". This suggests that DeepEL's performance represents a conservative estimate, with potentially higher actual effectiveness.

The second type of error involves expanding the scope of entity reference, denoted as \textit{Expand}. In these cases, DeepEL typically selects an answer that has an inclusion relationship with the correct entity. Conversely, the third error type involves narrowing the scope of entity reference, denoted as \textit{Narrow}. With this error type, DeepEL provides more specific and detailed answers than required.

The final error type is failing to provide an answer, denoted as \textit{Not give answer}. In these instances, DeepEL fails to select an answer despite the ground truth being correct and the candidate entity list containing the appropriate option.

Among these four error types, the latter three occur with similar frequency. This pattern emerges from a common cause: in DeepEL's chain of thought, these three errors typically arise when LLMs provide overly detailed or intrusive explanations of the current entity, ultimately influencing the final judgment.

\subsection{Case Study} 

In this section, we analyze DeepEL's error predictions in detail to better understand the underlying causes of its mistakes. Apart from errors stemming from inaccurate ground truth, the other three error types are well-documented and share similar characteristics. Within DeepEL's reasoning process, the initial entity descriptions provided by LLMs play a pivotal role in the framework's overall performance. When these descriptions are excessively detailed or broad, they can significantly influence subsequent judgment.

For example, as shown in Table 4, Row 3, when evaluating the entity "Sony", the LLM interprets it as 'Sony Corporation', highlighting its artistic achievements and numerous signed musicians. This comprehensive analysis increases the weighting for "Music", leading to the selection of "Sony Music" - a narrower referent than intended. Conversely, if the LLM interprets the entity more broadly, a type 2 error may occur, resulting in the selection of a candidate entity with a wider scope than appropriate. In extreme cases, if the LLM's misinterpretation prevents matching with any candidate entities, no answer is provided, resulting in a type 4 error.

\section{Conclusion}

In this study, we present the DeepEL framework, a comprehensive approach to entity linking that comprises three sequential stages. Initially, a pre-trained model generates a list of candidate entities based on descriptive inputs derived from LLMs. In the second stage, a multiple-choice disambiguation mechanism selects from the candidate set, effectively narrowing down potential entities. Finally, a self-validation phase systematically reassesses and refines preliminary selections, ensuring more accurate and contextually appropriate final outputs.

DeepEL deeply integrates LLMs' capabilities at each stage, fully leveraging their strengths in the entity linking task. Compared to existing models, DeepEL requires no fine-tuning, achieves more significant performance improvements, and demonstrates higher average accuracy. In addition, DeepEL shows strong robustness, achieving excellent results on parts of the dataset that are very difficult for other models. We believe that there is a lot of potential waiting to be explored in using LLMs for entity linking tasks.

\bibliographystyle{splncs04}
\bibliography{main}

\end{document}